\documentclass[runningheads]{llncs}

\usepackage[mobile]{eccv}
\usepackage[symbol]{footmisc}

\usepackage{eccvabbrv}

\usepackage{graphicx}
\usepackage{booktabs}

\usepackage[accsupp]{axessibility}  %

\usepackage[pagebackref,breaklinks,colorlinks,citecolor=eccvblue]{hyperref}

\usepackage{orcidlink}

\begin{document}

\title{Comp4D: LLM-Guided Compositional \\ 4D Scene Generation}

\titlerunning{Comp4D: Compositional 4D Scene Generation}

\author{Dejia Xu\inst{1*} \and
Hanwen Liang\inst{2*} \and
Neel P. Bhatt\inst{1} \and Hezhen Hu\inst{1} \and Hanxue Liang\inst{3} \and Konstantinos N. Plataniotis\inst{2} \and Zhangyang Wang\inst{1}}

\authorrunning{D. Xu, H. Liang, et al.}

\institute{University of Texas at Austin \and University of Toronto \and University of Cambridge}

\maketitle
\vspace{-4mm}
\footnotetext[1]{Equal contribution.}

\begin{figure}[h]
    \centering
    \includegraphics[width=\linewidth]{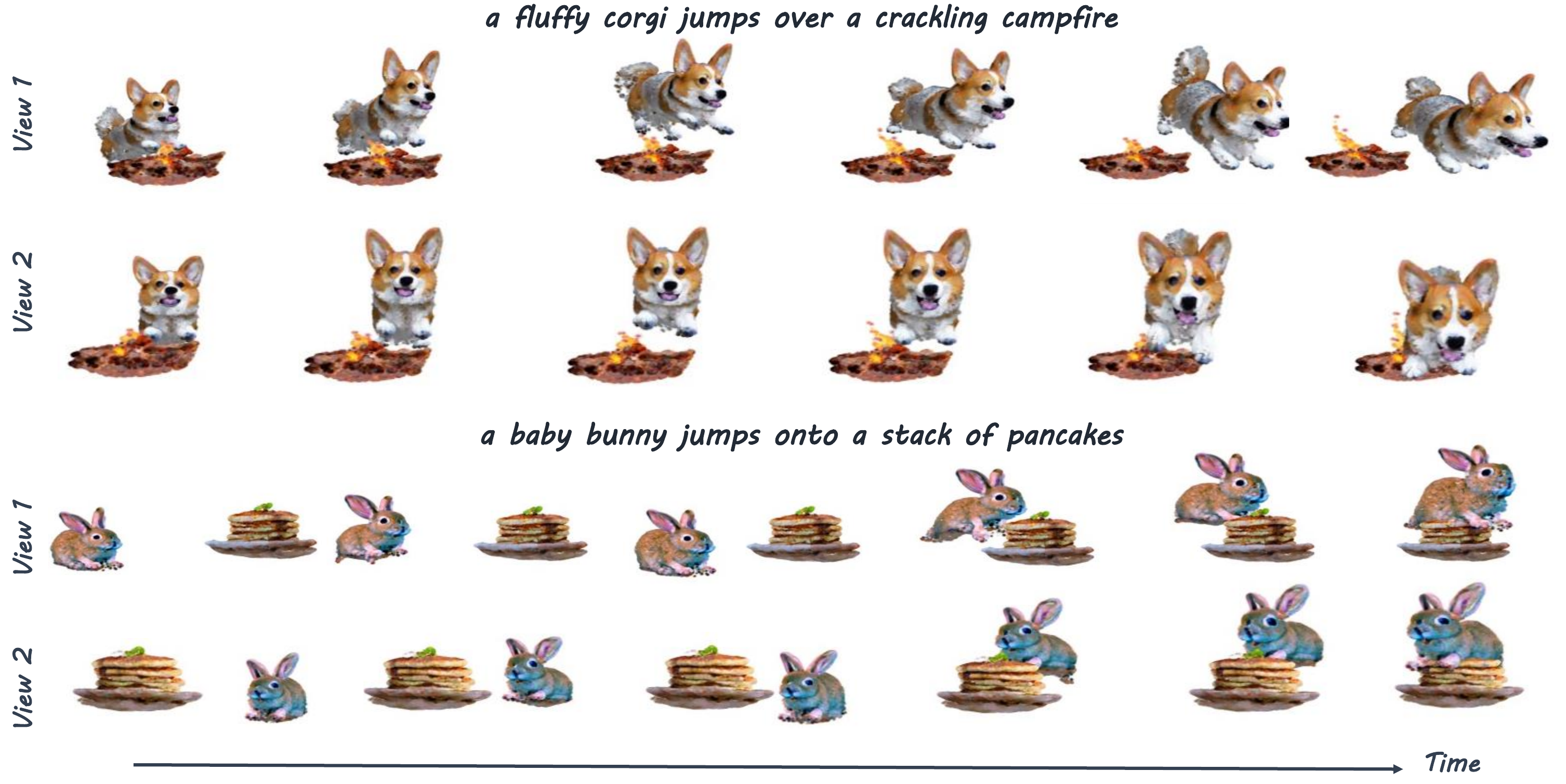}
\caption{We present Comp4D: compositional 4D scene synthesis from text input. Our model can render realistic images from various viewpoints and different timesteps.}
\label{fig:teaser}
\vspace{-8mm}
\end{figure}

\begin{abstract}
Recent advancements in diffusion models for 2D and 3D content creation have sparked a surge of interest in generating 4D content. However, the scarcity of 3D scene datasets constrains current methodologies to primarily object-centric generation. To overcome this limitation, we present Comp4D, a novel framework for compositional 4D scene generation. Unlike conventional methods that generate a singular 4D representation of the entire scene, Comp4D innovatively constructs each 4D object within the scene separately. Utilizing Large Language Models (LLMs), the framework begins by decomposing an input text prompt into distinct entities and maps out their trajectories. It then constructs the compositional 4D scene by accurately positioning these objects along their designated paths. To refine the scene, our method employs a compositional score distillation technique guided by the pre-defined trajectories, utilizing pre-trained diffusion models across text-to-image, text-to-video, and text-to-3D domains. Extensive experiments demonstrate our outstanding 4D content creation capability compared to prior arts, showcasing superior visual quality, motion fidelity, and enhanced object interactions. Project page: \url{https://vita-group.github.io/Comp4D}
\end{abstract}

\section{Introduction}
\label{sec:intro}

Recent advances in text-to-image diffusion models~\cite{imagen,nichol2021glide,ramesh2022hierarchical,rombach2022high} have revolutionized generative AI, simplifying digital content creation. Traditional pipelines, often cumbersome and requiring domain expertise, are being replaced by these generative models that bring complex ideas to life from simple text prompts. This innovation extends to the domain of 3D content creation, where score distillation techniques~\cite{poole2022dreamfusion,xu2022neurallift,wang2023score,shi2023mvdream,wang2023prolificdreamer,liu2023zero} leverage 2D diffusion models to generate 3D content. Yet, the focus remains predominantly on static 3D assets, reflecting the significant influence of text-to-image models.

While text-to-image diffusion models made significant strides in video generation, there is limited exploration into adapting these models for 4D content creation. The primary challenge is the absence of comprehensive 4D datasets, leading most research to focus on refining score distillation techniques to enhance novel view renderings of 4D representations. Concurrently, these works often rely on partial supervision signals including text prompts~\cite{ling2023align,mav3d,zheng2023unified,yin20234dgen,4dfy}, images~\cite{yin20234dgen,ren2023dreamgaussian4d,zhao2023animate124}, 3D models~\cite{yin20234dgen,zheng2023unified}, or monocular videos~\cite{yin20234dgen,jiang2023consistent4d,ren2023dreamgaussian4d}, to guide the generation process.

Despite notable advancements, current 4D content creation predominantly yields object-centric outputs. This limitation is mainly attributed to the scarcity of comprehensive scene-level 3D datasets. MVDream~\cite{shi2023mvdream} and Zero-123~\cite{liu2023zero,shi2023zero123++} trained on Objaverse~\cite{deitke2023objaverse} are widely adopted in 4D content creation pipelines~\cite{ling2023align,mav3d,zheng2023unified,yin20234dgen,ren2023dreamgaussian4d,zhao2023animate124,jiang2023consistent4d,4dfy}, which provide direct supervision on the multi-view renderings with geometry awareness. Compared to 2D diffusion models, these 3D-aware diffusion models greatly improve the 3D geometry quality in the generated content~\cite{liu2023zero,shi2023mvdream}.
However, their focus on object-centric generation persists, attributed to the reliance on the training data~\cite{deitke2023objaverse,deitke2023objaverse2}, which is comprised mostly of synthetic objects positioned at the world origin.
Notably, when an object's global movement is synchronized with the camera, it primarily exhibits local deformation in rendered images.
However, a complex 4D scene with multiple objects always exhibits not only individual local deformation but also inter-object global displacement.

\begin{figure}[t]
    \centering
    \includegraphics[width=\linewidth]{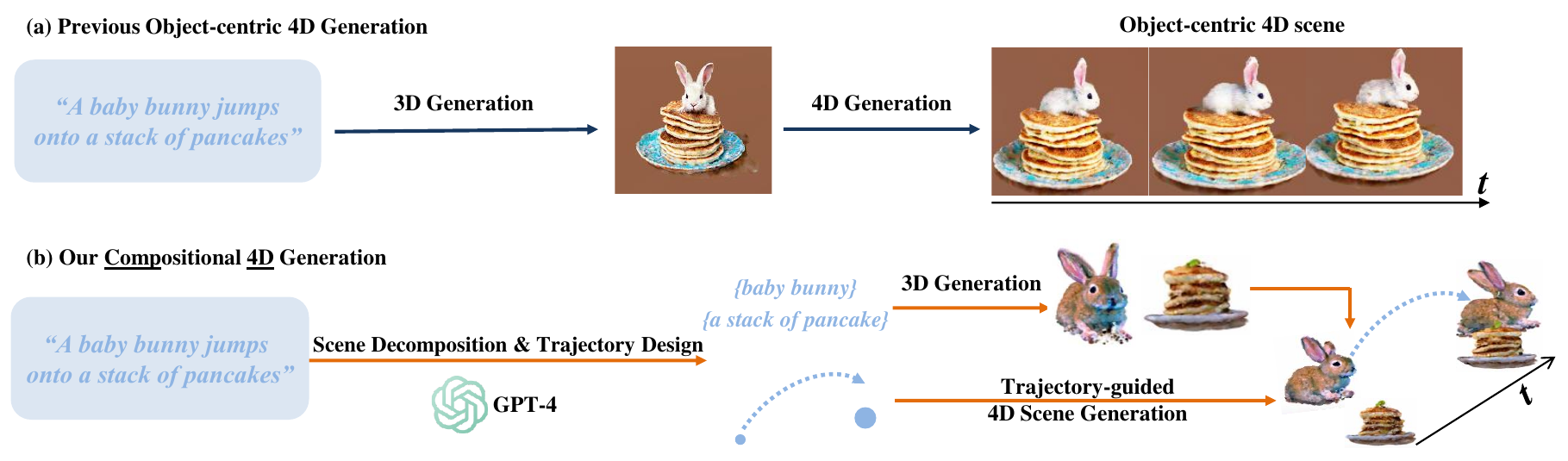}
    \caption{Compared with previous object-centric 4D generation pipelines, our \underline{Comp}ositional \underline{4D} Generation (\textbf{Comp4D}) framework integrates GPT-4 to decompose the scene and design proper trajectories, resulting in larger-scale movements and more realistic object interactions.%
    }
    \label{fig:task}
\end{figure}

\looseness=-1
Contrary to previous efforts concentrated on object-centric 4D objects~\cite{ling2023align,mav3d,zheng2023unified,yin20234dgen,ren2023dreamgaussian4d,zhao2023animate124,jiang2023consistent4d,4dfy}, our work extends the boundaries to the demanding task of constructing \textbf{compositional 4D scenes}, as illustrated in Fig.~\ref{fig:task}.
To overcome the prevalent object-centric constraint, our approach disentangles the compositional 4D scene generation process into two primary components: scene decomposition for constructing individual 3D assets, and motion generation with object interactions. 
The motion generation is further disentangled into two stages. First, a Large Language Model (LLM) is utilized to delineate global trajectories for each object, thereby alleviating the computational load on deformation modules by narrowing their focus to local deformations. 
Further, by formulating each object as deformable 3D Gaussians, we introduce a dynamic rendering mechanism to learn object local deformations. We selectively render the whole scene or single object during the optimization process. This strategy acts as a powerful augmentation to ensure motion fidelity of each object, especially in scenarios where object occlusion is prevalent as objects move.

\looseness=-1
The generation of our 4D scene is conducted through the following stages.
Given an input text prompt, we first leverage an LLM to extract entities and determine their attributes, such as scale. Following this, static 3D objects are individually constructed using 3D-aware diffusion models. 
Subsequently, the LLM determines the initial locations and the speed of the objects, and generates kinematics-based trajectories to model object global displacement. Finally, each object's deformation field is optimized via a novel compositional score distillation mechanism, with objects moving along the pre-defined trajectories.

Our key contributions can be summarized as follows:
\begin{itemize}
    \item We introduce \textbf{Comp4D}, a pioneering framework that achieves \underline{Comp}ositional \underline{4D} scene creation. By formulating 4D scene creation as the construction of individual 4D objects and their interactions, we overcome the object-centric constraint posed in previous methods.
    \item We propose to decompose object motion into global displacement and local deformation components. An LLM is tasked to design the global displacement of the object motion via kinematics templates, thereby offloading the 4D representation burden to concentrate solely on local deformation.
    \item Formulating each object as a set of deformable 3D Gaussians enables our 4D representation to flexibly switch between single-object and multi-object renderings. This design facilitates stable optimization of object motion even in the presence of potential occlusions.
    \item Extensive experiments compared to existing baselines demonstrate the superiority of our model in compositional 4D scene generation in terms of visual quality, motion realism, and object interaction.
\end{itemize}

\section{Related Works}
\label{sec:related}

\subsection{4D Content Creation}
Text-guided diffusion models have significantly advanced image and video generation. However, the scarcity of large text-annotated 3D training datasets constrains progress in 3D generative learning. To address this limitation, the score distillation technique~\cite{poole2022dreamfusion} is proposed for optimization-based text-to-3D generation.
Presently, score distillation has evolved into a flourishing research avenue for synthesizing static 3D objects, and some researchers have extended it to dynamic 4D scene generation.
Make-A-Video3D (MAV3D)~\cite{mav3d} is a pioneering work generating dynamic 4D scenes from text prompts. It uses NeRFs with HexPlane features for 4D representation. 4D-fy~\cite{4dfy} leverages a NeRF-based representation with a multi-resolution feature grid, combining supervision signals from images, videos, and 3D-aware diffusion models for text-to-4D synthesis. Dream-in-4D~\cite{zheng2023unified} introduces a novel text-to-4D synthesis scheme focusing on image-guided and personalized generation. 
Consistent4D~\cite{jiang2023consistent4d} tackles the task of video-to-4D generation with the help of RIFE~\cite{huang2022real} and a 2D super-resolution module.
Recently, with the advances in 3D Gaussians, AYG~\cite{ling2023align} proposes generating 4D scenes using dynamic 3D Gaussians, disentangling the 4D representation into static 3D Gaussian Splatting and a deformation field for modeling dynamics. 
4DGen~\cite{yin20234dgen} introduces a driving video to ground 4D content creation from images or text, providing added controllability in motion generation through reconstruction loss alongside score distillation. DreamGaussian4D~\cite{ren2023dreamgaussian4d} adopts mesh texture refining through video diffusion models to improve texture quality. However, existing works in 4D content creation focus on object-centric generation due to the underlying constraints of the 3D-aware diffusion model. In contrast, we make the first attempt to tackle the challenging compositional 4D scene generation task by decomposing the scene into object components.

\subsection{4D Scene Representation}
\looseness=-1
Building 4D representation in the scene allows for reconstructing and rendering novel images of objects under rigid and non-rigid motions from a single camera moving around the scene. 
D-NeRF~\cite{pumarola2021d} considers time as an additional input and divides the learning into two stages, \emph{i.e.,} one for encoding the scene into a canonical space and another to map this canonical representation into the deformed scene at a particular time. 
3D Cinemagraphy~\cite{li20233d} converts the input image into feature-based layered depth images, un-projects them to a feature point cloud, and animates it via lifted 3D scene flow.

Moreover, 3D Gaussian Splatting~(3D-GS)~\cite{kerbl20233d} has shown advantages in both effectiveness and efficiency, leading to multiple directions to model temporal dynamics.
Katsumata~\emph{et al.}~\cite{katsumata2023efficient} and 4DGS~\cite{wu20234d} define scales, positions, and rotations as functions of time while leaving other time-invariant properties of the static 3D Gaussians unchanged.
Another direction involves directly extending 3D Gaussians to 4D with temporal slicing~\cite{yang2023real,duan20244d}.
There are also works leveraging a separate function to model the dynamic distribution of attributes' deformation for 3D Gaussians~\cite{lin2023gaussian,li2023spacetime}.
In this work, we adopt 3D Gaussians for our 3D content representation and use an additional Multi Layer Perceptron (MLP) to deform each set of 3D Gaussians for ease of optimization. This disentangled 4D representation allows us first to construct the static scene and then focus on generating the object's deformation.

\subsection{Grounding and Reasoning from Large Language Models}
LLMs have emerged as a natural tool for performing reasoning tasks and enabling implementation in the real world. Recently developed LLMs have demonstrated great few-shot performance across a variety of tasks through in-context learning \cite{chen2022improving}. These models can be prompted with a question and trained using several input and output examples to solve a problem through reasoning \cite{rajani2019explain}.

A popular approach to improving the reasoning capabilities of LLMs is to fine-tune models on domain-specific tasks \cite{yang2023fine}. Moreover, researchers have investigated methods for grounding LLMs by incorporating structured knowledge bases, ontologies, and external databases to provide contextual understanding and improve reasoning capabilities. For instance, efforts have been made to integrate symbolic reasoning techniques with LLMs, enabling them to infer logical relationships and perform deductive reasoning tasks \cite{LiZZYLZWYZHCG22, Lu2023MultimodalPP, brohan2023can}.

Moreover, recent studies have explored techniques for incorporating multimodal information, such as images and videos, to enhance contextual understanding and improve the robustness of language models, paving the way for more effective applications in various domains \cite{seff2023motionlm, aerospace10090770}.

Specifically, LLMs have recently been used for generating trajectories in robotics applications. For instance, in \cite{kwon2023language,bucker2023latte}, dense trajectories were generated for a manipulator by an LLM in a zero-shot manner. The demonstration confirms the potential of LLMs to act as trajectory generators. Given that LLMs can reason about the high-level task at hand, they can generate trajectories that match the overall goal.

\section{Method}

\begin{figure}[t]
    \centering
    \includegraphics[width=\linewidth]{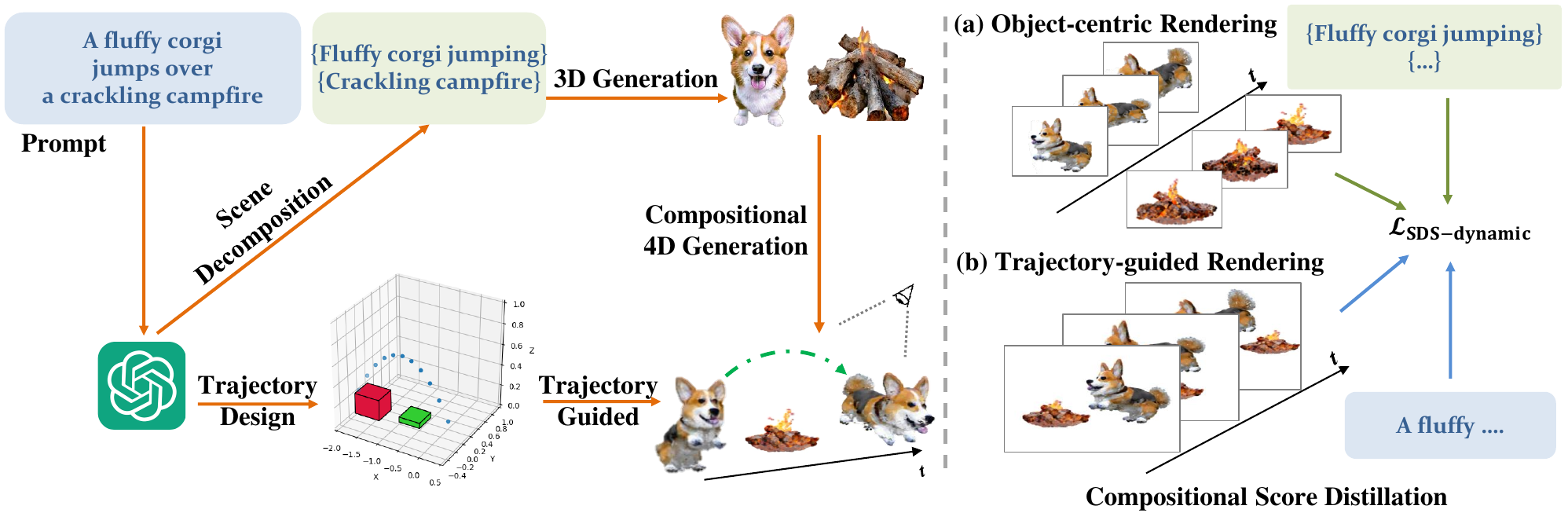}
    \caption{An overview of our proposed {Comp4D} method. Given an input text, first, we use LLM for scene decomposition to obtain multiple individual 3D objects. Subsequently, we adopt LLM to design the object trajectories which guide the displacements of objects in the optimization of compositional 4D scene. Thanks to the compositional 4D representation implemented with 3D Gaussians, in each iteration of the compositional score distillation process, we are able to switch between object-centric rendering and trajectory-guided rendering flexibly.}
    \label{fig:method}
\end{figure}

\looseness=-1
In this section, we illustrate the components of our proposed method in detail (Fig.~\ref{fig:method}). We start by introducing some preliminaries (Sec.~\ref{sec:prelim}) on 3D Gaussians and score distillation sampling. Then we introduce our compositional 4D scene representation (Sec.~\ref{sec:4drep}). We later discuss how we leverage LLM for scale assignment and trajectory design (Sec.~\ref{sec:llm}). Finally, we illustrate our compositional score distillation pipeline involving multiple diffusion models (Sec.~\ref{sec:sds}).

\subsection{Preliminaries}
\label{sec:prelim}

\paragraph{\textbf{3D Gaussian Splatting}}
3D Gaussian Splatting~\cite{kerbl20233d}, which we refer to as 3D-GS, parameterizes a 3D scene as a set of 3D Gaussians. Each Gaussian is defined with a center position $\mu$, covariance $\Sigma$, opacity $\alpha$, and color $c$ modeled by spherical harmonics. Unlike implicit representation methods such as
NeRF~\cite{mildenhall2021nerf}, which renders images based on volumetric rendering, 3D-GS renders images through a tile-based rasterization operation, achieving real-time rendering speed.

Starting from a set of points randomly initialized in the unit sphere, each point is designated a 3D Gaussian, which can be queried as follows:
\begin{equation}
\footnotesize
    {G}(x)=e^{-\frac{1}{2}(x-\mu)^{T}\Sigma^{-1}(x-\mu)},
\label{eq:gaussian}
\end{equation}
where $x$ is an arbitrary position in the 3D scene. During the rendering process, the 3D
Gaussians $G(x)$ are first transformed to 2D Gaussians $G^{'}(x)$ on the image plane. Then a tile-based rasterizer is designed to
efficiently sort the 2D Gaussians and employ $\alpha$-blending:
\begin{equation}
\begin{aligned}
\footnotesize
    C(r)=\sum_{i\in N}c_{i}\sigma_{i}\prod_{j=1}^{i-1}(1-\sigma_{j}),~~~\sigma_{i}=\alpha_{i}G'(r),
    \label{eq:rasterization}
\end{aligned}
\end{equation}
where $r$ is the queried pixel position, $N$ denotes the number of sorted 2D Gaussians associated with the queried pixel, while $c_{i}$ and $\alpha_{i}$
denote the color and opacity of the $i$-th Gaussian. In our experiments, we empirically simplify the color of Gaussians to diffuse color for the sake of efficient training.

\paragraph{\textbf{Score Distillation Sampling}} 
Current methodologies for text-to-3D or 4D generation typically involve iterative optimization of a scene representation with supervisory signals from pre-trained diffusion models~\cite{poole2022dreamfusion, wang2023prolificdreamer}.
Initially, rendering of the 3D or 4D scene is acquired in the form of an image or sequence of images. 
Random noise is added to the rendered images, and a pre-trained diffusion model is employed to de-noise the images. 
The estimated gradient from this process is utilized to update the 3D or 4D representations.  
Specifically, employing a 3D representation parameterized by $\theta$ and a rendering method $g$, the rendered images are generated as $x = g(\theta)$. 
To align the rendered image $x$ with samples obtained from the diffusion model $\phi$, the diffusion model employs a score function $\hat{\epsilon}_{\phi}(x_{t};y,t)$ to predict a noise map $\hat{\epsilon}$, given the noise level $t$, noisy input $x_{t}$ and text embeddings $y$. 
By evaluating the difference between the Gaussian noise ${\epsilon}$ added to the rendered images $x$ and the predicted noise $\hat{\epsilon}$, this score function updates the parameter $\theta$ with gradient formulated as:
\begin{equation}
\begin{aligned}
\footnotesize
    \nabla_{\theta}\mathcal{L}_{SDS}(\phi,x=g(\theta))=w(t)(\hat{\epsilon}_{\phi}(x_{t};y,t)-\epsilon)\frac{\partial{x}}{\partial{\theta}},
    \label{eq:sds}
\end{aligned}
\end{equation}
where $w(t)$ is a weighting function. Using Score Distillation Sampling (SDS) for text-to-4D generation requires coordinated guidance to achieve realistic outcomes in terms of appearance, 3D structure, and motion~\cite{4dfy}. This often involves the utilization of hybrid SDS, which combines both image-based and video-based diffusion models~\cite{ling2023align}.

\subsection{Compositional 4D Representation}
\label{sec:4drep}

We build our 4D representation through the composition of individual objects.
For each object, we utilize a set of static 3D Gaussians along with an MLP-based deformation network.
The MLP network takes in $(x, y, z, t)$ coordinates as input and outputs the 3D deformation of point locations. Following the best practice in previous works~\cite{tancik2020fourier,mildenhall2021nerf}, the input coordinates are processed with positional encoding as a 32-dimensional vector to enable high-frequency feature learning. 
This decomposed design of architecture supports decoupled learning of the static attributes of an object (\textit{e.g.} geometry and texture) and the motion information.
We start our training stage by optimizing the static 3D Gaussian attributes. Once they converge, we introduce the deformation field and freeze the static 3D Gaussian attributes (i.e. covariance, opacity and color) to stabilize the training process.

Naively optimizing the deformation field leads to unpleasant results. This is mainly due to the fact that the MLP modulates each point location individually, ignoring the overall rigidity of the object. 
Similar to AYG~\cite{ling2023align}, we adopt rigidity constraints to ensure that the deformation of each Gaussian is consistent with its k-nearest neighbors,
\begin{equation}
    \mathcal{L}_\text{rigidity}(x) = \frac{1}{k}\sum_{i=1}^k || \Delta_x - \Delta_{x_{NN_i}} ||.
\end{equation}

Moreover, to avoid flickering motion, we introduce additional regularization loss components that penalize sudden changes in the acceleration of each 3D Gaussian,
\begin{equation}
    \mathcal{L}_\text{acc}(x, t) = ||\Delta_{x, t} + \Delta_{x, t + 2} - 2 \Delta_{x, {t+1}}||.
\label{eq:acc}
\end{equation}

Thanks to the explicit nature of 3D Gaussians, at rendering time, we can selectively render a single object or multiple objects. This provides the freedom to better supervise the motion fidelity of the individual objects as well as their interactions. However, since the objects are represented separately as 3D Gaussians, no explicit constraint can prevent the objects from intersecting with each other. Once the objects are overlapping, the rendered image will show collapsed shapes resulting in unstable gradients from score distillation. To this end, we draw inspiration from CG3D~\cite{vilesov2023cg3d} to incorporate physics-based contact loss that avoids the collision of multiple objects. For one object, we ensure the contact angle $\theta_j$ for each 3D Gaussian with mean $\mu_j$ to be acute:
\begin{equation}
\begin{aligned}
    \theta_j & = (c - \mu_i) \cdot (\mu_j - \mu_i), \\
    \mathcal{L}_\text{contact} &= -\theta_j [\theta_j < 0],
\label{eq:contact}
\end{aligned}
\end{equation}
where $\mu_i$ refers to the mean of the Gaussian in the other object that is closest to $\mu_j$, and c denotes the center of the current object.

\begin{figure}[t]
    \centering
    \includegraphics[width=\linewidth]{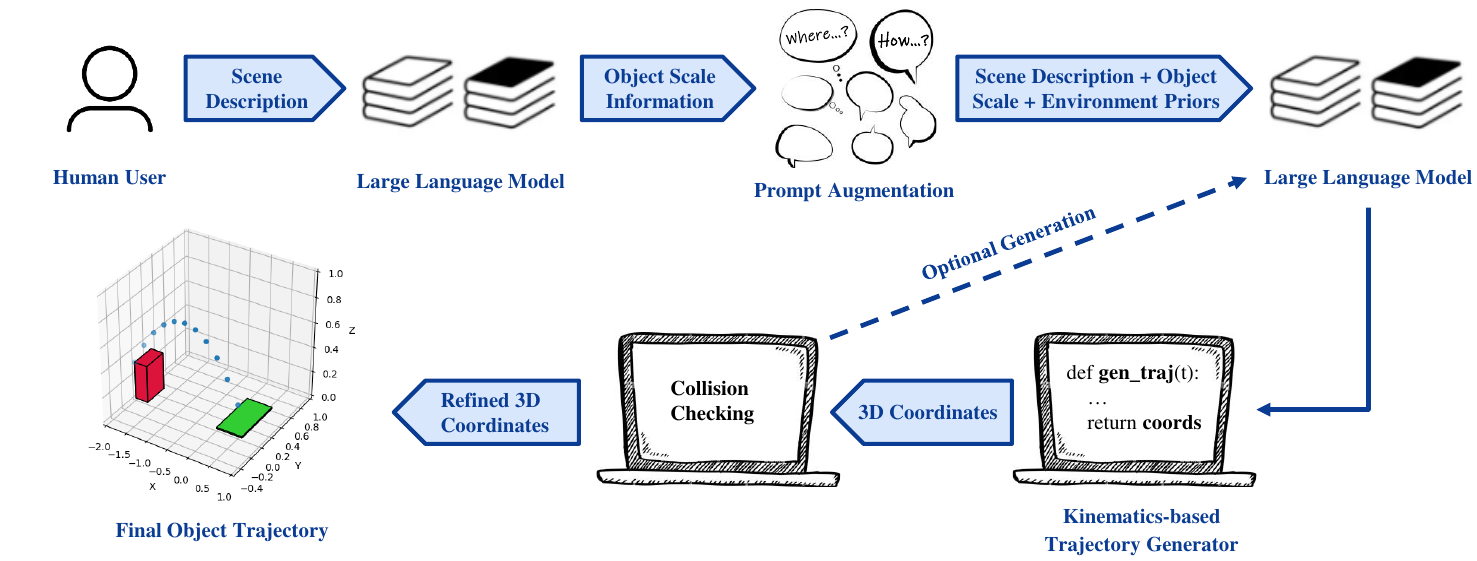}
    \caption{The overall pipeline for object trajectory generation. First, a scene description is provided by a human user as a prompt to an LLM which yields relative object scale information required for rendering. Subsequently, the language model is prompted with this information along with environmental constraints and tasked to return a function that takes timestep as an input and returns the corresponding object's 3D position governed by kinematic equations. After the collection of a set of positions, collision checking is performed to clip the trajectory where the first collision occurs. Optionally, premature collisions can be mitigated by re-querying the LLM for an improved function.}
    \label{fig: llm-traj-gen-pipeline}
\end{figure}

\subsection{LLM Guided Scale Assignment and Trajectory Design}
\label{sec:llm}

We decompose the object's motion into a global displacement and local deformation. We offload the task of designing global displacement to an LLM and, therefore, reduce the workload for the 4D representation and let the distillation models focus on producing realistic local deformation.
We illustrate the overall pipeline for object trajectory generation in Fig.~\ref{fig: llm-traj-gen-pipeline}.

\paragraph{\textbf{Scale Assignment}}
As we start the 4D scene synthesis by generating the static 3D assets, the resulting objects are generally of similar scales. This is mainly because the 3D-aware diffusion models are trained on synthetic 3D objects that are normalized to unit scale.
Therefore, determining the appropriate scale of each asset becomes crucial as it enables a realistic and reasonable composition of the scene. 
Recent studies~\cite{li2023comprehensive,bubeck2023sparks} show that GPT-4 demonstrates remarkable ability in reasoning with commonsense knowledge. 
Therefore, we directly prompt the language model to make reasonable assumptions of the relative scale of the objects. 
We adaptively resize our static 3D assets to corresponding scales to ensure their composition resulting in a realistic scene.

\paragraph{\textbf{Trajectory Design Through Kinematics Templates}}
We leverage the reasoning capability of GPT-4 to select physics-based formulas that govern the motion of objects. These formulas consist of kinematics-based equations such as those governing parabolic motion for objects acting as projectiles under the influence of gravity (Eq. \ref{eq:kinematics}).

\begin{equation} \label{eq:kinematics}
    \Delta x = v_it + \frac{1}{2}at^2.
\end{equation}

Furthermore, GPT-4 adeptly determines a sensible initial location and velocity for the moving object, ensuring that the generated trajectory aligns with the scene description. 
To streamline the optimization of object deformation, we instruct the model to assume that one object is positioned at the world origin and solely predict the trajectory of relative motion. 
Note that during the training of our model, both objects are allowed to deform.

\paragraph{\textbf{Optional Trajectory Refinement via Collision Checking}}
Despite curated prompt engineering, we observe that the trajectories proposed by GPT-4 can be imperfect occasionally. This is mainly due to its ignorance of the actual shape of 3D Gaussians and only taking the rough bounding box size into the reasoning process. 
As a result, the produced trajectory tends to contain unexpected collisions involving objects overlapping with each other at the end of the trajectory.

Therefore, we introduce an optional trajectory refinement step as a workaround to ensure the quality of the generated trajectory. We, therefore, uniformly sample points along the trajectory and obtain the object location at corresponding timestamps. We rotate the objects accordingly such that the canonical orientation of the object faces toward the next location sampled from the trajectory. More details about object placement are presented in the following section. After we obtain the object placement at each timestamp, we utilize Eq.~\ref{eq:contact} to determine if there is an ongoing collision. We then truncate the trajectory at the first collision to avoid the intersection of objects during rendering. If the final trajectory is too short to perform reasonable object motion, we can optionally choose to re-generate the entire trajectory by prompting GPT-4 again.

\subsection{4D Scene Optimization via Compositional Score Distillation}
\label{sec:sds}

\paragraph{\textbf{Static 3D Construction}}

We start our 4D scene generation by constructing each static 3D object. In order to ensure both photo-realism of texture and consistent geometry, we draw inspiration from Magic123~\cite{qian2023magic123} and 4d-fy~\cite{4dfy} to incorporate the joint distillation of image diffusion~\cite{rombach2022high} and a 3D-aware diffusion model~\cite{shi2023mvdream}. Specifically, we adopt the weighted combination of two sets of score distillation losses. Given a batch of rendered image $x$ and text embeddings $y$, the loss function is formulated as follows,
\small
\begin{equation}
    \nabla_{\theta}\mathcal{L}_\text{SDS-static}(x, y) = \{\omega_\text{sd-static}(\epsilon_\text{sd}(x_{t1}; y, t_1) - \epsilon_1) + \omega_\text{mv}(\epsilon_\text{mv}(x_{t2}; y, t_2) - \epsilon_2)\}\frac{\partial{x}}{\partial{\theta}},
\end{equation}
where $\omega_\text{sd-static}$ and $\omega_\text{mv}$ are coefficients for the score distillation loss of Stable Diffusion~\cite{rombach2022high} and MVDream~\cite{shi2023mvdream}.

\paragraph{\textbf{Trajectory-Guided Scene Optimization}}
With the help of the pre-defined trajectories, the objects can have their locations specified easily. By sampling from the trajectory function, $F(\cdot)$, we can obtain the object locations at arbitrary timesteps $t_i$. Objects are rotated accordingly such that their canonical orientation faces toward the next location along the trajectory $\overrightarrow{R_i} = (F(t_{i+1}) - F(t_{i}))$. Thanks to MVDream~\cite{shi2023mvdream}, which generates objects in their canonical orientation, our static stage produces objects facing the same direction (\textit{e.g.} $\overrightarrow{R_0} = (1, 0, 0)$), ensuring that our rotation strategy will produce objects moving towards their head direction. Given normalized head direction $A=\frac{\overrightarrow{R_0}}{||\overrightarrow{R_0}||}$ and $B=\frac{\overrightarrow{\bar{R_i}}}{\overrightarrow{\bar{R_i}}}$, the axis of rotation $v$ is obtained as $v=A \times B$. The angle of rotation $\theta$ is determined by $\cos(\theta)= A \cdot B$. We then obtain the skew-symmetric matrix $\mathbf{K}$ as follows,
\begin{equation}
    \mathbf{K} = \begin{bmatrix}
    0 & -v_z & v_y \\ v_z & 0 & -v_x \\ -v_y & v_x & 0
    \end{bmatrix},
\end{equation}
which is then used in Rodrigues' rotation formula to obtain the final rotation matrix $\mathbf{R}$,
\begin{equation}
    \mathbf{R} = \mathbf{I} + (\sin \theta) \mathbf{K} + (1 - \cos \theta) \mathbf{K}^2.
\end{equation}
Thanks to the pre-processed trajectory, our framework supports distilling objects with long-range motion and multi-concept interactions, which is difficult to achieve using previous baselines. 

Besides global motion, we utilize a deformation MLP for each set of 3D Gaussian for local motion learning.
To better learn the deformation field, we leverage a text-to-video diffusion model~\cite{zeroscope} to formulate the score distillation loss. Similar to distilling a static 3D object via an image diffusion model, score distillation via a video diffusion model ensures that the renderings at consecutive frames form a natural video aligned with the text prompt. 
As observed in previous works~\cite{4dfy,ling2023align}, image diffusion models usually generate a more realistic appearance compared to video diffusion models. Therefore, we jointly distill the score from image diffusion on individual frames to ensure texture quality. The loss function can be formulated as follows,

\small
\begin{equation}
\label{eq:sds_dyn}
    \nabla_{\theta}\mathcal{L}_\text{SDS-dyn}(x, y) = \{\omega_\text{sd-dyn}(\epsilon_\text{sd}(x_{t1}; y, t_1) - \epsilon_1) + \omega_\text{vid}(\epsilon_\text{vid}(x_{t2}; y, t_2) - \epsilon_2)\}\frac{\partial{x}}{\partial{\theta}},
\end{equation}
where $x$ is the generated image sequence and $y$ is the text prompt. $\omega_\text{sd-dyn}$ and $\omega_\text{vid}$ are coefficients for the score distillation loss from image- and video-based diffusion models.

\paragraph{\textbf{Object-Centric Motion Learning}}
Thanks to our compositional design, our framework supports arbitrary rendering combinations. This means that in each iteration, we have the flexibility to choose whether to render each object or not. This provides us with great freedom in rendering the scene with a diverse appearance. Such diversity provides rich augmentations that are essential for the stable optimization of score distillation loss. Otherwise, the occlusion of multiple objects in the same scene will make score distillation loss ineffective in ensuring reasonable motion of each object. In the single-object motion learning phase, the text prompt is modified by removing inactive entities, which avoids disturbing learning the deformation of the current object.
To this end, we leverage GPT-4 to decompose the scene prompt into individual prompts for each entity. Consequently, when rendering a single object, we supervise the renderings with score distillation losses (Eq.~\ref{eq:sds_dyn}) conditioned on the corresponding text prompt.

\begin{figure}[!htpb]
    \centering
    \includegraphics[width=0.99\linewidth]{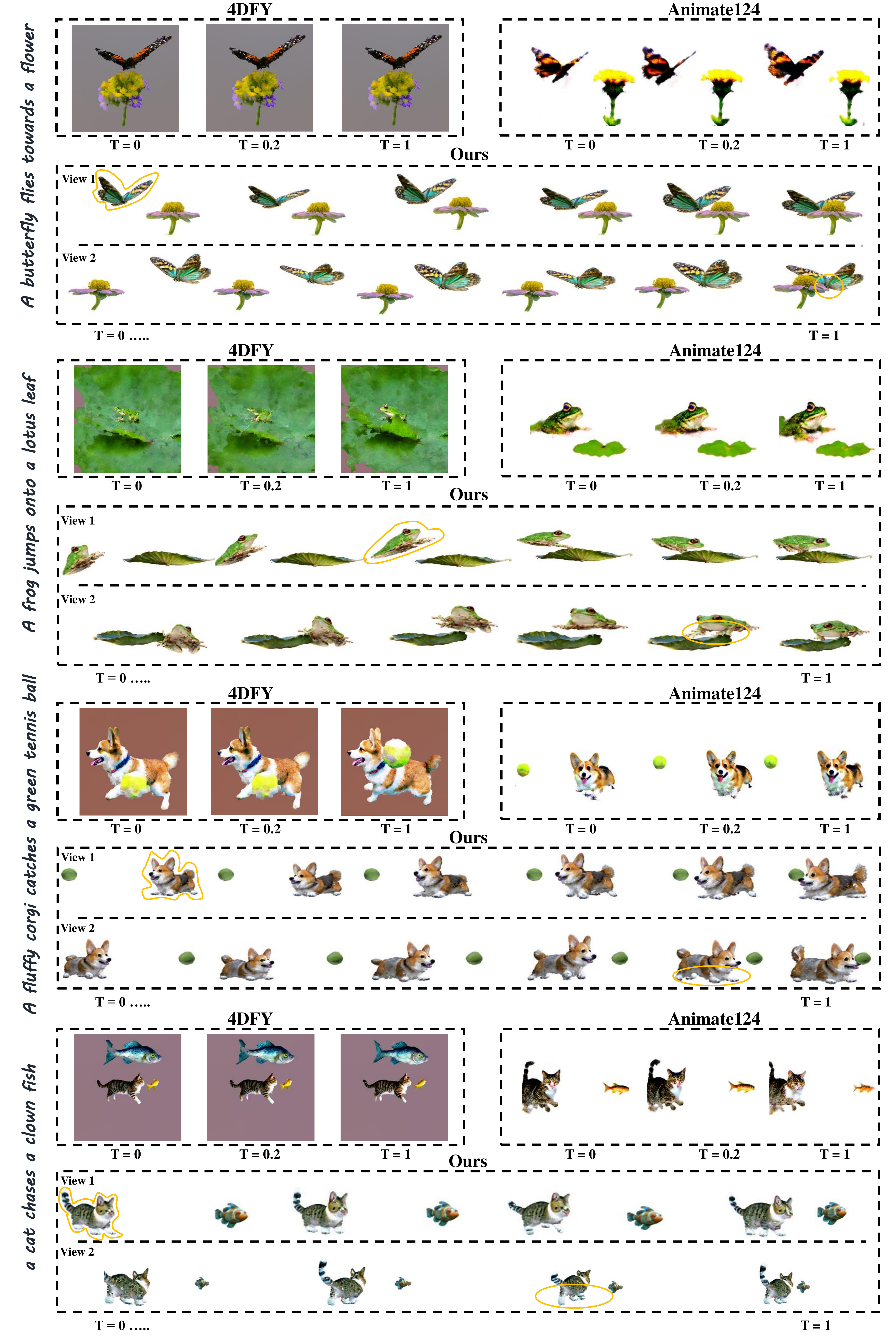}
    \caption{Comparison with previous object-centric 4D generation pipelines. Our Comp4D framework generates compositional 4D scenes with more realistic motion and object interactions. %
    }
    \label{fig:exp}
\end{figure}

\section{Experiments}

\subsection{Implementation Details}
Similar to Magic123~\cite{qian2023magic123} and 4dfy~\cite{4dfy}, we first generate the static 3D objects via NeRF representation using joint score distillation from MVDream~\cite{shi2023mvdream} and Stable Diffusion 2.1~\cite{rombach2022high}. After obtaining the static objects, we convert them to point clouds which are consecutively used to initialize 3D Gaussians. 
Our full model utilizes point clouds containing 60,000 colored points. We preprocess 60 nearest neighbors for each 3D Gaussian in order to speed up the calculation of $\mathcal{L}_\text{rigidity}$.
In the dynamic optimization stage, we randomly assign training iterations to adopt single-object rendering (with a probability of 0.2) or compositional rendering (with a probability of 0.8). We train for 3,000 iterations in the dynamic stage with a learning rate of 1e-4 for the deformation MLP. In each iteration, we render 16 frames via uniformly sampled timesteps. We use the frozen video diffusion model, Zeroscope~\cite{zeroscope}, in our experiments.
To improve the 2D appearance, we also randomly sample 4 frames out of 16 rendered frames for image score distillation, where the Stable Diffusion 2.1~\cite{rombach2022high} is used as the image diffusion model.
In our experiments, we compare with two open-source baseline text-to-4D generation methods, 4dfy~\cite{4dfy} and animate124~\cite{zhao2023animate124}.

\subsection{Main Results}

\begin{figure}[t]
    \centering
    \includegraphics[width=0.85\textwidth]{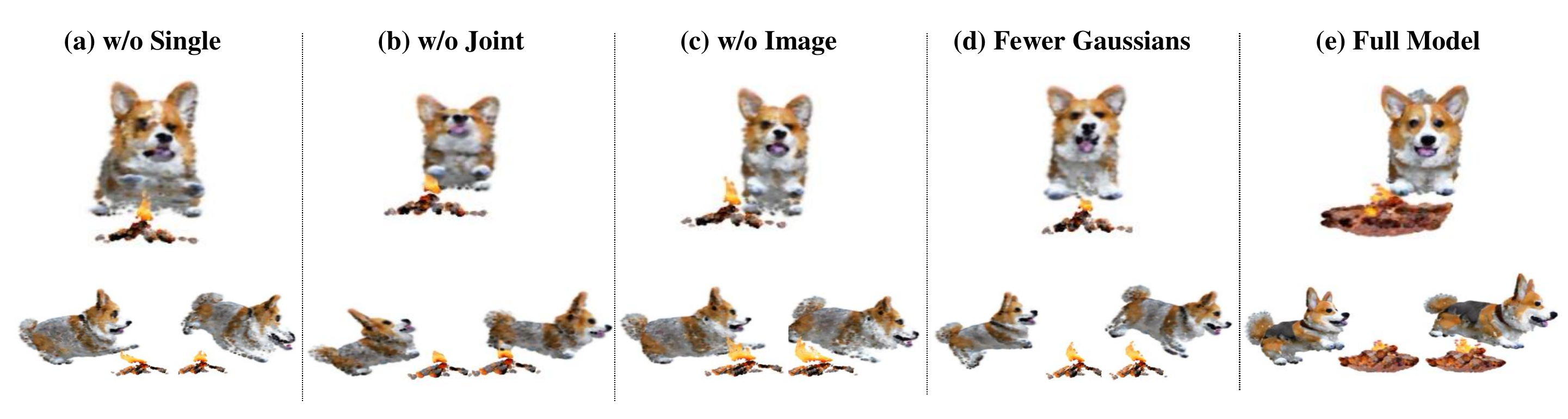}
    \caption{Ablation studies on the proposed components. The first row shows the front view. The second row shows the side view. Note that (a)-(d) are conducted using fewer number of Gaussians.} %
    \label{fig:ablation}
\end{figure}

\paragraph{\textbf{Metrics}}
In the absence of ground truth for unsupervised text-to-4D scene generation, we seek help from non-reference quality-assessment models for images and videos in the wild. Q-Align~\cite{wu2023q} is a recently proposed large multi-modal model fine-tuned from mPLUG-Owl2~\cite{ye2023mplug} using in-the-wild image and video quality assessment datasets. QAlign provides quality assessment functionality for images and videos in terms of aesthetics and quality. It has achieved state-of-the-art performance in alignment with human ratings on existing quality assessment benchmarks. The output scores are in the range of 1 (worst) to 5 (best).
As shown in Tab.~\ref{tab:comparison}, we report the average scores on four views ($0^\circ$,$90^\circ$,$180^\circ$, $270^\circ$) of our test samples and observe that our method outperforms existing methods in all metrics by a large margin.

\paragraph{\textbf{Qualitative Results}}
In Fig.~\ref{fig:exp}, we provide a detailed visualization of generated scenes at different timesteps from various views.
With the same text prompt, we compare our method with 4dfy~\cite{4dfy} and animate124~\cite{zhao2023animate124}.
For these prior works, we show the scenes from one view at timestamps of 0, 0.2, and 1s. For our method, we show two views with uniformly sampled timestamps from 0 to 1s.
As shown in the image, our framework excels in generating lifelike single objects with expansive motions while enhancing fidelity in object interactions.
As indicated by the yellow contours in Fig.~\ref{fig:exp}, we can observe the distinct flapping of butterfly wings, dynamic changes in body shape as the frog jumps, and variations in body contours of the corgi and cat as they run and leap. 
Comparatively, the object motion in baseline methods is minimal.
Going through the timeline, we can find that the objects move according to the pre-generated trajectory and display more frequent and realistic interactions. As illustrated in the yellow circles in the first two examples, the butterfly settles on the petal, and the frog stretches out its legs on the lotus leaf.
In comparison, baseline methods tend to generate static objects centered at the origin with texture flickering to simulate the movement of the objects. %

\paragraph{\textbf{Resolution and Speed}}
4dfy~\cite{4dfy} conducts the video score distillation stage at a resolution of $160 \times 288$. 
Similarly, animate124~\cite{zhao2023animate124} performs score distillation at a resolution of $80 \times 144$ due to NeRF's expensive rendering cost.
Contrarily, our method can render video at a resolution of $320 \times 576$ during score distillation which aligns with the training resolution of video diffusion~\cite{zeroscope} and facilitates superior motion generation. 
At inference time, thanks to the efficient Gaussian representation, our 4D scene representation renders at 70 FPS at $320 \times 576$. In comparison, 4dfy and animate124 render at around 4 FPS.

\subsection{Ablation Studies}

We evaluate the effectiveness of all the proposed components in Fig.~\ref{fig:ablation} and Tab.~\ref{tab:ablation}. 
To save computation costs, we utilize 3D Gaussians containing 20,000 points during the ablation study.
In Fig.~\ref{fig:ablation}(a), ``w/o Single'' refers to the variant without object-centric rendering. We observe the worst object geometry possibly due to the occlusions occurring in the optimization process. ``w/o Joint'' in Fig.~\ref{fig:ablation}(b) denotes that we only perform object-centric rendering without rendering two objects altogether in the same scene. The final results exhibit decreased motion and reduced interactions. In Fig.~\ref{fig:ablation}(c), we remove the SDS loss from image diffusion and only distill with video diffusion. Consequently, we observe that objects appear to have poor textures compared to (d) where image diffusion SDS loss is included. In Fig.~\ref{fig:ablation}(d), the model training and losses are kept the same as the full model (e), except that the number of 3D Gaussians we generate in the static stage is fewer. As shown in the figure, using fewer Gaussians results in less detailed texture and less realistic geometry. In summary, our full model (e) delivers the best results both quantitatively and qualitatively.

\begin{table}[t]
    \centering
    \caption{Quantitative comparison against baseline methods.}
    \begin{tabular}{l|cccc}
     \toprule
        Metric & 4dfy~\cite{4dfy} & Animate124~\cite{zhao2023animate124} & Ours\\
        \hline
        QAlign-img-quality $\uparrow$& 2.031 & 1.434 & \bf{2.931} \\ 
        QAlign-img-aesthetic $\uparrow$& 1.767 & 1.484 & \bf{2.190} \\ 
        QAlign-vid-quality $\uparrow$& 2.465 & 1.948 & \bf{3.367} \\ 
        QAlign-vid-aesthetic $\uparrow$& 1.973 & 1.654 & \bf{2.461} \\ 
    \midrule
        Rendering FPS $\uparrow$ & 4 & 4 & \textbf{70} \\
    \bottomrule
    \end{tabular}
    \label{tab:comparison}
\end{table}

\begin{table}[t]
    \centering
    \caption{Ablation studies on our proposed components.}
    \begin{tabular}{l|c|c|c|c|c}
    \toprule
        Metric & w/o Single & w/o Joint & w/o Image & Fewer Gaussians & Full Model\\
        \hline
        QAlign-img-quality $\uparrow$ & 1.8252 & 1.9893 & 1.8613 & 1.9131 & \textbf{2.4785} \\ 
        QAlign-img-aesthetic $\uparrow$  & 1.6455 &  1.8789 & 1.7715 & 1.8301  & \textbf{1.9004} \\
        QAlign-vid-quality $\uparrow$ & 2.4082 & 2.4102 & 2.3926 & {2.7285} & \textbf{2.9023}\\ 
        QAlign-vid-aesthetic $\uparrow$ & 1.9062 & 1.9512 & 1.9014 & {2.0039} & \textbf{2.1621} \\ 
    \bottomrule
    \end{tabular}
    \label{tab:ablation}
\end{table}

\section{Limitations and Future Work}
Despite the exciting results produced by Comp4D, our framework still has some limitations. First, we are leveraging the zero-shot ability of GPT-4, which can be further improved if the language model is fine-tuned to generate a more precise trajectory and more complex motion. 
Second, the motion we generate is constrained by the ability of video diffusion models (\textit{e.g.} realism and length). 
Moving forward, we will explore generating longer and more complex motions for more practical 4D content creation.

\section{Conclusion}
In this work, we present Comp4D, a novel framework for generating compositional 4D scenes from text input. With the help of GPT-4, we decompose scene generation into the creation of individual objects as well as their interactions. Given a compositional scene description, we first leverage GPT-4 to generate object prompts for the independent creation of 3D objects. Subsequently, it is tasked to design the trajectory for the moving objects. This predefined trajectory then guides the compositional score distillation process, which optimizes a composable 4D representation comprising deformable 3D Gaussians for each object. Our experiments demonstrate that Comp4D significantly surpasses existing text-to-4d generation methods in terms of visual quality, motion fidelity, and object interactions.

\bibliographystyle{splncs04}
\bibliography{main}

\begin{thebibliography}{10}
\providecommand{\url}[1]{\texttt{#1}}
\providecommand{\urlprefix}{URL }
\providecommand{\doi}[1]{https://doi.org/#1}

\bibitem{zeroscope}
Zeroscope. \url{https://huggingface.co/cerspense/zeroscope_v2_576w} (2023)

\bibitem{4dfy}
Bahmani, S., Skorokhodov, I., Rong, V., Wetzstein, G., Guibas, L., Wonka, P., Tulyakov, S., Park, J.J., Tagliasacchi, A., Lindell, D.B.: 4d-fy: Text-to-4d generation using hybrid score distillation sampling. arXiv preprint arXiv:2311.17984  (2023)

\bibitem{bubeck2023sparks}
Bubeck, S., Chandrasekaran, V., Eldan, R., Gehrke, J., Horvitz, E., Kamar, E., Lee, P., Lee, Y.T., Li, Y., Lundberg, S., et~al.: Sparks of artificial general intelligence: Early experiments with gpt-4. arXiv preprint arXiv:2303.12712  (2023)

\bibitem{bucker2023latte}
Bucker, A., Figueredo, L., Haddadin, S., Kapoor, A., Ma, S., Vemprala, S., Bonatti, R.: Latte: Language trajectory transformer. In: 2023 IEEE International Conference on Robotics and Automation (ICRA). pp. 7287--7294. IEEE (2023)

\bibitem{chen2022improving}
Chen, M., Du, J., Pasunuru, R., Mihaylov, T., Iyer, S., Stoyanov, V., Kozareva, Z.: Improving in-context few-shot learning via self-supervised training. arXiv preprint arXiv:2205.01703  (2022)

\bibitem{deitke2023objaverse2}
Deitke, M., Liu, R., Wallingford, M., Ngo, H., Michel, O., Kusupati, A., Fan, A., Laforte, C., Voleti, V., Gadre, S.Y., et~al.: Objaverse-xl: A universe of 10m+ 3d objects. arXiv preprint arXiv:2307.05663  (2023)

\bibitem{deitke2023objaverse}
Deitke, M., Schwenk, D., Salvador, J., Weihs, L., Michel, O., VanderBilt, E., Schmidt, L., Ehsani, K., Kembhavi, A., Farhadi, A.: Objaverse: A universe of annotated 3d objects. In: Proceedings of the IEEE/CVF Conference on Computer Vision and Pattern Recognition. pp. 13142--13153 (2023)

\bibitem{duan20244d}
Duan, Y., Wei, F., Dai, Q., He, Y., Chen, W., Chen, B.: 4d gaussian splatting: Towards efficient novel view synthesis for dynamic scenes. arXiv preprint arXiv:2402.03307  (2024)

\bibitem{huang2022real}
Huang, Z., Zhang, T., Heng, W., Shi, B., Zhou, S.: Real-time intermediate flow estimation for video frame interpolation. In: European Conference on Computer Vision. pp. 624--642. Springer (2022)

\bibitem{brohan2023can}
Ichter, B., Brohan, A., Chebotar, Y., Finn, C., Hausman, K., Herzog, A., Ho, D., Ibarz, J., Irpan, A., Jang, E., Julian, R., Kalashnikov, D., Levine, S., Lu, Y., Parada, C., Rao, K., Sermanet, P., Toshev, A., Vanhoucke, V., Xia, F., Xiao, T., Xu, P., Yan, M., Brown, N., Ahn, M., Cortes, O., Sievers, N., Tan, C., Xu, S., Reyes, D., Rettinghouse, J., Quiambao, J., Pastor, P., Luu, L., Lee, K., Kuang, Y., Jesmonth, S., Joshi, N.J., Jeffrey, K., Ruano, R.J., Hsu, J., Gopalakrishnan, K., David, B., Zeng, A., Fu, C.K.: Do as {I} can, not as {I} say: Grounding language in robotic affordances. In: Conference on Robot Learning. Proceedings of Machine Learning Research, vol.~205, pp. 287--318. PLMR, Auckland, New Zealand (2022)

\bibitem{jiang2023consistent4d}
Jiang, Y., Zhang, L., Gao, J., Hu, W., Yao, Y.: Consistent4d: Consistent 360 $\{$$\backslash$deg$\}$ dynamic object generation from monocular video. arXiv preprint arXiv:2311.02848  (2023)

\bibitem{katsumata2023efficient}
Katsumata, K., Vo, D.M., Nakayama, H.: An efficient 3d gaussian representation for monocular/multi-view dynamic scenes. arXiv  (2023)

\bibitem{kerbl20233d}
Kerbl, B., Kopanas, G., Leimk{\"u}hler, T., Drettakis, G.: 3d gaussian splatting for real-time radiance field rendering. ACM Transactions on Graphics (ToG)  \textbf{42}(4),  1--14 (2023)

\bibitem{kwon2023language}
Kwon, T., Di~Palo, N., Johns, E.: Language models as zero-shot trajectory generators. In: 2nd Workshop on Language and Robot Learning: Language as Grounding (2023)

\bibitem{LiZZYLZWYZHCG22}
Li, L.H., Zhang, P., Zhang, H., Yang, J., Li, C., Zhong, Y., Wang, L., Yuan, L., Zhang, L., Hwang, J., Chang, K., Gao, J.: Grounded language-image pre-training. In: Conference on Computer Vision and Pattern Recognition. pp. 10955--10965. {IEEE}, New Orleans, LA, USA (2022)

\bibitem{li20233d}
Li, X., Cao, Z., Sun, H., Zhang, J., Xian, K., Lin, G.: 3d cinemagraphy from a single image. In: CVPR. pp. 4595--4605 (2023)

\bibitem{li2023comprehensive}
Li, Y., Wang, L., Hu, B., Chen, X., Zhong, W., Lyu, C., Zhang, M.: A comprehensive evaluation of gpt-4v on knowledge-intensive visual question answering. arXiv preprint arXiv:2311.07536  (2023)

\bibitem{li2023spacetime}
Li, Z., Chen, Z., Li, Z., Xu, Y.: Spacetime gaussian feature splatting for real-time dynamic view synthesis. arXiv preprint arXiv:2312.16812  (2023)

\bibitem{lin2023gaussian}
Lin, Y., Dai, Z., Zhu, S., Yao, Y.: Gaussian-flow: 4d reconstruction with dynamic 3d gaussian particle. arXiv preprint arXiv:2312.03431  (2023)

\bibitem{ling2023align}
Ling, H., Kim, S.W., Torralba, A., Fidler, S., Kreis, K.: Align your gaussians: Text-to-4d with dynamic 3d gaussians and composed diffusion models. arXiv preprint arXiv:2312.13763  (2023)

\bibitem{liu2023zero}
Liu, R., Wu, R., Van~Hoorick, B., Tokmakov, P., Zakharov, S., Vondrick, C.: Zero-1-to-3: Zero-shot one image to 3d object. arXiv preprint arXiv:2303.11328  (2023)

\bibitem{Lu2023MultimodalPP}
Lu, Y., Lu, P., Chen, Z., Zhu, W., Wang, X.E., Wang, W.Y.: Multimodal procedural planning via dual text-image prompting (2023)

\bibitem{mildenhall2021nerf}
Mildenhall, B., Srinivasan, P.P., Tancik, M., Barron, J.T., Ramamoorthi, R., Ng, R.: Nerf: Representing scenes as neural radiance fields for view synthesis. Communications of the ACM  \textbf{65}(1),  99--106 (2021)

\bibitem{nichol2021glide}
Nichol, A., Dhariwal, P., Ramesh, A., Shyam, P., Mishkin, P., McGrew, B., Sutskever, I., Chen, M.: Glide: Towards photorealistic image generation and editing with text-guided diffusion models. arXiv preprint arXiv:2112.10741  (2021)

\bibitem{poole2022dreamfusion}
Poole, B., Jain, A., Barron, J.T., Mildenhall, B.: Dreamfusion: Text-to-3d using 2d diffusion. arXiv preprint arXiv:2209.14988  (2022)

\bibitem{pumarola2021d}
Pumarola, A., Corona, E., Pons-Moll, G., Moreno-Noguer, F.: D-nerf: Neural radiance fields for dynamic scenes. In: Proceedings of the IEEE/CVF Conference on Computer Vision and Pattern Recognition. pp. 10318--10327 (2021)

\bibitem{qian2023magic123}
Qian, G., Mai, J., Hamdi, A., Ren, J., Siarohin, A., Li, B., Lee, H.Y., Skorokhodov, I., Wonka, P., Tulyakov, S., et~al.: Magic123: One image to high-quality 3d object generation using both 2d and 3d diffusion priors. arXiv preprint arXiv:2306.17843  (2023)

\bibitem{rajani2019explain}
Rajani, N.F., McCann, B., Xiong, C., Socher, R.: Explain yourself! leveraging language models for commonsense reasoning. arXiv preprint arXiv:1906.02361  (2019)

\bibitem{ramesh2022hierarchical}
Ramesh, A., Dhariwal, P., Nichol, A., Chu, C., Chen, M.: Hierarchical text-conditional image generation with clip latents. arXiv preprint arXiv:2204.06125  (2022)

\bibitem{ren2023dreamgaussian4d}
Ren, J., Pan, L., Tang, J., Zhang, C., Cao, A., Zeng, G., Liu, Z.: Dreamgaussian4d: Generative 4d gaussian splatting. arXiv preprint arXiv:2312.17142  (2023)

\bibitem{rombach2022high}
Rombach, R., Blattmann, A., Lorenz, D., Esser, P., Ommer, B.: High-resolution image synthesis with latent diffusion models. In: Proceedings of the IEEE/CVF conference on computer vision and pattern recognition. pp. 10684--10695 (2022)

\bibitem{imagen}
Saharia, C., Chan, W., Saxena, S., Li, L., Whang, J., Denton, E., Ghasemipour, S.K.S., Gontijo-Lopes, R., Ayan, B.K., Salimans, T., Ho, J., Fleet, D.J., Norouzi, M.: Photorealistic text-to-image diffusion models with deep language understanding. In: Oh, A.H., Agarwal, A., Belgrave, D., Cho, K. (eds.) Advances in Neural Information Processing Systems (2022), \url{https://openreview.net/forum?id=08Yk-n5l2Al}

\bibitem{seff2023motionlm}
Seff, A., Cera, B., Chen, D., Ng, M., Zhou, A., Nayakanti, N., Refaat, K.S., Al{-}Rfou, R., Sapp, B.: Motionlm: Multi-agent motion forecasting as language modeling. arXiv preprint arXiv:2309.16534  (2023)

\bibitem{shi2023zero123++}
Shi, R., Chen, H., Zhang, Z., Liu, M., Xu, C., Wei, X., Chen, L., Zeng, C., Su, H.: Zero123++: a single image to consistent multi-view diffusion base model. arXiv preprint arXiv:2310.15110  (2023)

\bibitem{shi2023mvdream}
Shi, Y., Wang, P., Ye, J., Long, M., Li, K., Yang, X.: Mvdream: Multi-view diffusion for 3d generation. arXiv preprint arXiv:2308.16512  (2023)

\bibitem{mav3d}
Singer, U., Sheynin, S., Polyak, A., Ashual, O., Makarov, I., Kokkinos, F., Goyal, N., Vedaldi, A., Parikh, D., Johnson, J., et~al.: Text-to-4d dynamic scene generation. arXiv preprint arXiv:2301.11280  (2023)

\bibitem{tancik2020fourier}
Tancik, M., Srinivasan, P., Mildenhall, B., Fridovich-Keil, S., Raghavan, N., Singhal, U., Ramamoorthi, R., Barron, J., Ng, R.: Fourier features let networks learn high frequency functions in low dimensional domains. Advances in Neural Information Processing Systems  \textbf{33},  7537--7547 (2020)

\bibitem{aerospace10090770}
Tikayat~Ray, A., Bhat, A.P., White, R.T., Nguyen, V.M., Pinon~Fischer, O.J., Mavris, D.N.: Examining the potential of generative language models for aviation safety analysis: Case study and insights using the aviation safety reporting system (asrs). Aerospace  \textbf{10}(9) (2023)

\bibitem{vilesov2023cg3d}
Vilesov, A., Chari, P., Kadambi, A.: Cg3d: Compositional generation for text-to-3d via gaussian splatting. arXiv preprint arXiv:2311.17907  (2023)

\bibitem{wang2023score}
Wang, H., Du, X., Li, J., Yeh, R.A., Shakhnarovich, G.: Score jacobian chaining: Lifting pretrained 2d diffusion models for 3d generation. In: Proceedings of the IEEE/CVF Conference on Computer Vision and Pattern Recognition. pp. 12619--12629 (2023)

\bibitem{wang2023prolificdreamer}
Wang, Z., Lu, C., Wang, Y., Bao, F., Li, C., Su, H., Zhu, J.: Prolificdreamer: High-fidelity and diverse text-to-3d generation with variational score distillation. arXiv preprint arXiv:2305.16213  (2023)

\bibitem{wu20234d}
Wu, G., Yi, T., Fang, J., Xie, L., Zhang, X., Wei, W., Liu, W., Tian, Q., Wang, X.: 4d gaussian splatting for real-time dynamic scene rendering. arXiv preprint arXiv:2310.08528  (2023)

\bibitem{wu2023q}
Wu, H., Zhang, Z., Zhang, W., Chen, C., Liao, L., Li, C., Gao, Y., Wang, A., Zhang, E., Sun, W., et~al.: Q-align: Teaching lmms for visual scoring via discrete text-defined levels. arXiv preprint arXiv:2312.17090  (2023)

\bibitem{xu2022neurallift}
Xu, D., Jiang, Y., Wang, P., Fan, Z., Wang, Y., Wang, Z.: Neurallift-360: Lifting an in-the-wild 2d photo to a 3d object with 360 $\{$$\backslash$deg$\}$ views. arXiv preprint arXiv:2211.16431  (2022)

\bibitem{yang2023fine}
Yang, Y., Bhatt, N.P., Ingebrand, T., Ward, W., Carr, S., Wang, Z., Topcu, U.: Fine-tuning language models using formal methods feedback. arXiv preprint arXiv:2310.18239  (2023)

\bibitem{yang2023real}
Yang, Z., Yang, H., Pan, Z., Zhu, X., Zhang, L.: Real-time photorealistic dynamic scene representation and rendering with 4d gaussian splatting. arXiv preprint arXiv:2310.10642  (2023)

\bibitem{ye2023mplug}
Ye, Q., Xu, H., Ye, J., Yan, M., Liu, H., Qian, Q., Zhang, J., Huang, F., Zhou, J.: mplug-owl2: Revolutionizing multi-modal large language model with modality collaboration. arXiv preprint arXiv:2311.04257  (2023)

\bibitem{yin20234dgen}
Yin, Y., Xu, D., Wang, Z., Zhao, Y., Wei, Y.: 4dgen: Grounded 4d content generation with spatial-temporal consistency. arXiv preprint arXiv:2312.17225  (2023)

\bibitem{zhao2023animate124}
Zhao, Y., Yan, Z., Xie, E., Hong, L., Li, Z., Lee, G.H.: Animate124: Animating one image to 4d dynamic scene. arXiv preprint arXiv:2311.14603  (2023)

\bibitem{zheng2023unified}
Zheng, Y., Li, X., Nagano, K., Liu, S., Hilliges, O., De~Mello, S.: A unified approach for text-and image-guided 4d scene generation. arXiv preprint arXiv:2311.16854  (2023)

\end{thebibliography}
\end{document}


\title{LLM-Guided Compositional 4D Scene Generation}

\titlerunning{Abbreviated paper title}

\author{First Author\inst{1}\orcidlink{0000-1111-2222-3333} \and
Second Author\inst{2,3}\orcidlink{1111-2222-3333-4444} \and
Third Author\inst{3}\orcidlink{2222--3333-4444-5555}}

\authorrunning{F.~Author et al.}

\institute{Princeton University, Princeton NJ 08544, USA \and
Springer Heidelberg, Tiergartenstr.~17, 69121 Heidelberg, Germany
\email{lncs@springer.com}\\
\url{http://www.springer.com/gp/computer-science/lncs} \and
ABC Institute, Rupert-Karls-University Heidelberg, Heidelberg, Germany\\
\email{\{abc,lncs\}@uni-heidelberg.de}}

\maketitle

\section{Additional Implementation Details}
Our overall loss function can be summarized as follows,
\begin{equation}
    \mathcal{L} =  \mathcal{L}_\text{SDS-static} +  \mathcal{L}_\text{SDS-dynamic} + \mathcal{L}_\text{reg},
\end{equation}
where $\mathcal{L}_\text{reg}$ refers to regularizations:
\begin{equation}
    \mathcal{L}_\text{reg} = \mathcal{L}_\text{contact} + \omega_1\mathcal{L}_\text{acc} + \omega_2\mathcal{L}_\text{rigidity}.
\end{equation}
$\omega_1$ and $\omega_2$ are weighting coefficients, set to $1e-4$ and $1e3$, respectively. For MVDream~\cite{shi2023mvdream}, we use negative prompt as their default configuration, ``ugly, bad anatomy, blurry, pixelated obscure, unnatural colors, poor lighting, dull, and unclear, cropped, lowres, low quality, artifacts, duplicate, morbid, mutilated, poorly drawn face, deformed, dehydrated, bad proportions''. For Zeroscope~\cite{zeroscope}, we use negative prompts that avoid static generations, ``static, low motion, static statue, not moving, no motion, text, watermark, copyright, blurry, nsfw''.

\section{Experiment Settings}
For baseline methods 4dfy~\cite{4dfy} and animate-124~\cite{zhao2023animate124}, we use the codebase released by the authors. For 4dfy~\cite{4dfy}, since it is a text-to-4D generation method, we use the same scene text description when generating 4D results. For animate-124~\cite{zhao2023animate124}, due to its image-to-4D nature, we use the static assets produced by our method to obtain the initial image for animate-124. Specifically, we place the static assets at the $t=0$ position in the trajectory and render a static image that does not contain occlusions. Then, we follow the workflow of animate-124 first to perform textual inversion and then the following static and dynamic stages.

\section{Additional Results}
\textbf{We provide additional results and comparisons in the video.}

\bibliographystyle{splncs04}
\bibliography{main}